\documentclass[letterpaper, 10 pt, conference]{ieeeconf}  

\IEEEoverridecommandlockouts                              

\usepackage{graphicx}
\usepackage{color}
\usepackage{amsfonts}
\usepackage{amsmath}
\usepackage{algorithm}
\usepackage{algpseudocode}
\usepackage{url}
\usepackage{tabularx}
\usepackage{multirow} 
\usepackage{booktabs} 
\usepackage{xcolor}

\title{\LARGE \bf
Transformer-Based Spatio-Temporal Association of Apple Fruitlets
}

\author{Harry Freeman and George Kantor
\thanks{H. Freeman and G. Kantor are with Carnegie Mellon University, Pittsburgh PA, USA \texttt{\{hfreeman, kantor\}@cs.cmu.edu}}
}

\begin{document}
\maketitle
\thispagestyle{empty}
\pagestyle{empty}

\begin{abstract}
In this paper, we present a transformer-based method to spatio-temporally associate apple fruitlets in stereo-images collected on different days and from different camera poses. State-of-the-art association methods in agriculture are dedicated towards matching larger crops using either high-resolution point clouds or temporally stable features, which are both difficult to obtain for smaller fruit in the field. To address these challenges, we propose a transformer-based architecture that encodes the shape and position of each fruitlet, and propagates and refines these features through a series of transformer encoder layers with alternating self and cross-attention. We demonstrate that our method is able to achieve an F1-score of 92.4\% on data collected in a commercial apple orchard and outperforms all baselines and ablations.
\end{abstract}


\section{Introduction}

The global food supply is constantly under increasing pressure as a result of climate change, population growth, and increased labor shortages. To keep up with demand, agriculturalists are turning to computer vision-based systems that can automate a variety of laborious and time-intensive tasks such as 
harvesting~\cite{harvest_mangaonkar2022fruit}, 
pruning~\cite{pruning_silwal2021bumblebee}, 
counting~\cite{counting_freeman20233d}, 
and crop 
modeling~\cite{modelling_kelly2023target}.
These automated solutions not only improve efficiency, but also help mitigate the challenges posed by labor shortages and increasing food demand, ensuring that critical agricultural tasks can be performed reliably at scale.

One particularly important but challenging task to automate is monitoring the growth and development of individual plants and fruits. Monitoring plant and fruit growth is important because it enables agricultural specialists to make more informed real-time crop management decisions and helps with downstream tasks such as 
phenotyping~\cite{ pheno_watt2020phenotyping},
disease management~\cite{dm_li2021plant}, 
and yield prediction~\cite{yield_he2022fruit}.
Apple growers, for example, track the growth rates of apple fruitlets\footnote{A fruitlet is a young apple formed on the tree shortly after pollination} to determine when to spray chemical thinner on their trees to prevent them from developing a pattern of alternative year bearing, ultimately allowing them to maintain a more consistent yearly harvest~\cite{greene2013development}. The current method of recording growth rates involves using a digital caliper to manually measure the sizes of the fruitlets in approximately 105 clusters\footnote{A cluster is a group of fruitlets that grow out of the same bud} per varietal, with each cluster typically containing five to six fruits. This is performed twice per thinning application, usually spread over four days, with the first measurement taken three to four days after application, and the second recorded seven to eight days after application. Computer vision-based systems could significantly reduce the labor required to size hundreds to thousands of fruits, potentially improving accuracy, speed, and cost-efficiency. Enhancing these factors is essential for scaling food production to meet growing global demand while mitigating the impacts of labor shortages and climate variability on agricultural productivity.

\begin{figure}[!t]
    \centering
    \includegraphics[width=\linewidth]{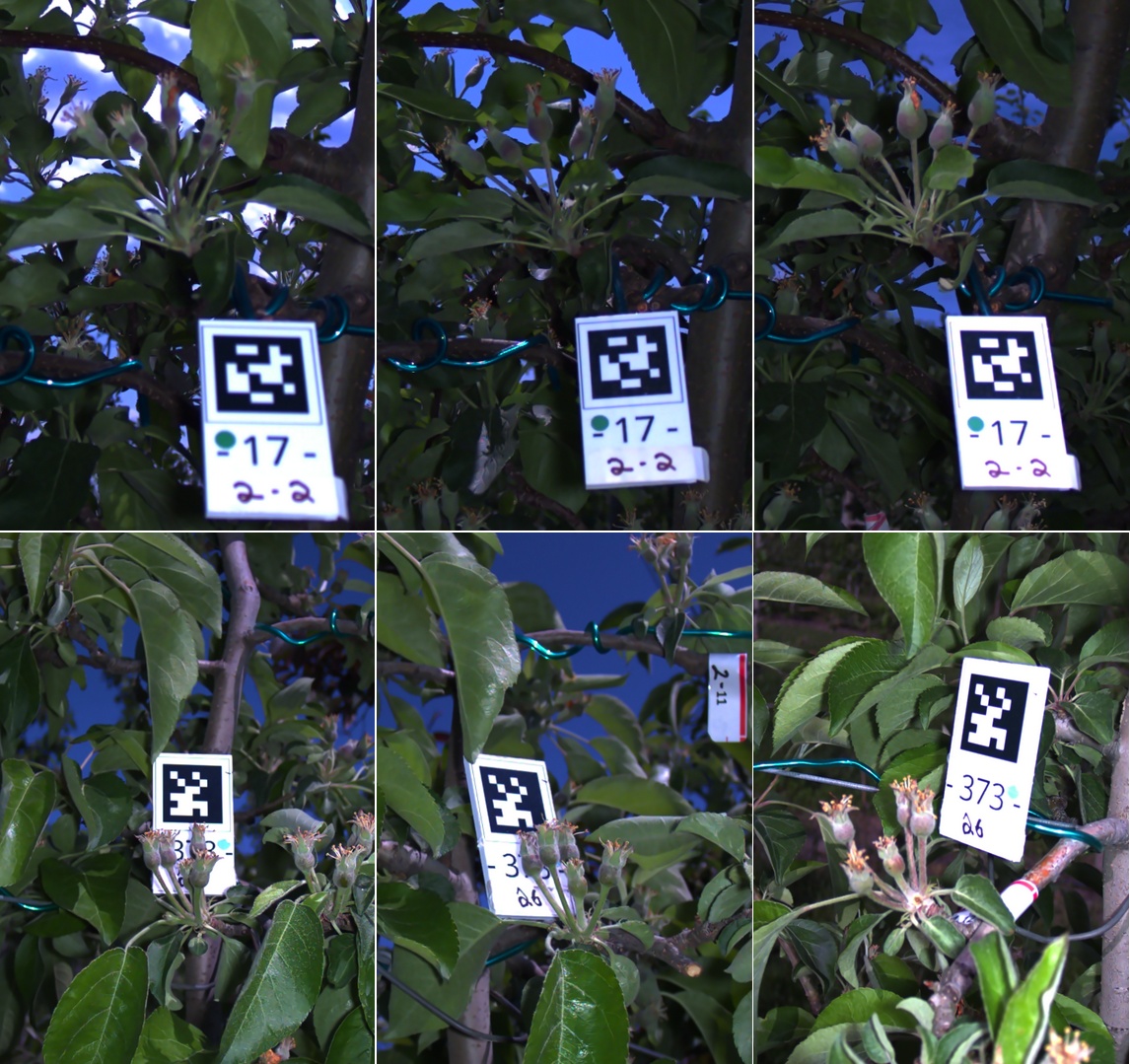}
    \vspace{-18pt}
    \caption{Top: Example of fruit growth over four day period between sizes. Left, middle, and right are day 1, day 3, and day 5 with average diameters 5.5mm, 7.9mm, and 9.7mm respectively. Bottom: Images of same fruitlet cluster taken from different camera poses which often occurs in the field.}
    \label{fig:cluster_example}
    \vspace{-15pt}
\end{figure}

In order to successfully automate growth monitoring, the fruit must be spatio-temporally associated. This is challenging in the field from a computer vision-based perspective, especially for smaller crops. This is because they are more difficult to detect consistently, are easily occluded, and exhibit significant variations in visual appearance and relative position as they grow. In the case of apple fruitlets~(Fig.~\ref{fig:cluster_example}), the fruits can be as small as 6mm in diameter when initially sized, can fall off or change position, and can nearly double in size over a period of four days~\cite{greene2013development}. They also undergo significant shape and color changes, further complicating association. Additionally, accurate sizing requires close-up images that minimize occlusions, which means camera poses may vary across days~(Fig.~\ref{fig:cluster_example}), as neither a human nor a robot can reliably capture structurally similar close-up images compared to images captured when driving down a row of crops~\cite{riccardi}. 

Although previous methods have focused on spatio-temporal association of plants and fruits in agriculture, they mainly target larger crops and rely on either high-precision laser scans collected in controlled lab environments~\cite{heiwolt2023statistical, pandey2024spatio} or temporally stable features~\cite{lobefarospatio}, which are difficult to obtain for small fruit in the field. Additionally, some approaches require multi-image registration~\cite{lobefaro}, which can be slow when sizing hundreds of fruits and prone to inaccuracies in leafy and dynamic environments.

To overcome these limitations, we present a transformer-based~\cite{transformer} approach for spatio-temporally associating apple fruitlets. Using single stereo image pairs collected in the field across days, our method encodes the shape and spatial geometry of the fruit, which are then matched using a transformer encoder architecture with alternating self- and cross-attention layers. We validate our approach against baselines and ablations, and demonstrate its generalizability by testing it on a dataset of different crops and modalities. Our primary contributions are:
\vspace{-5pt}
\begin{itemize}
\item A transformer-based approach to spatio-temporally associate apple fruitlets, leveraging shape and spatial geometry information.
\item An association method that relies solely on single stereo image pairs, eliminating the need for high-precision laser scanners or temporally stable features.
\item Evaluation on data collected in a commercial apple orchard and other datasets to demonstrate generalizability.
\end{itemize}

\section{Related Work}
Recent research has explored spatio-temporal association in agriculture. For large rows of crops, 4D association methods have been introduced at whole-plot scales, leveraging the assumptions of static scenes during data collection and the structural consistency of crop rows throughout the 
growing season~\cite{field_0, field_2, field_4}.
These methods struggle to adapt well when imaging individual plants and fruits, as the underlying assumptions break down due to greater variability in the structure of the local scene over time. 

Several works have also focused on the temporal registration of individual plant components on a single stem, many of which skeletonize the crop and associate the different skeletal components~\cite{skel_0, skel_1, skel_2, pandey2024spatio}. Skeletonization techniques are difficult to apply to apple fruitlets because of their elliptical shape and levels of occlusion. Other methods include using self-organizing maps~\cite{som} and fitting statistical models~\cite{ssm}. However, the point clouds of the plants were obtained using a high-precision laser scanner on a single isolated plant in a controlled laboratory environment. These techniques would not work on incomplete point clouds resulting from occlusions and missed detections, which are often experienced in the field. 
Recently, Lobefaro \textit{et al.} \cite{lobefaro} introduced a method to 4D associate features of plant point clouds using RGB-D images, and later extended their work to deformably register a reference map to a newly collected sequence of images taken at a later point in time~\cite{lobefarospatio}. However, they use multiple cameras, and their method relies on plant stems as temporally stable features, which are not present in our use case.

More similar to our work, Hondo \textit{et al.} \cite{fixed_apple} size and track growth rates of apples over time using a deep learning-based approach. The images are collected using a fixed camera, and association is naively performed by comparing the positions of segmented apples. This would not extend to measuring growth rates of apple fruitlets due to their dynamic nature and number needed to be sized.
Riccardi \textit{et al.}~\cite{riccardi} present a system to track and temporally register strawberries. They propose a histogram-based descriptor that encodes the spatial relationship between a fruit and its neighbors. However, the descriptor is not SO(3) invariant, making its performance highly dependent on accurate initial scene alignment, achieved with 
RANSAC~\cite{ransac}. 
In their approach, this alignment is more straightforward due to data being collected by a robot driving down a row of crops with a high-precision laser scanner. In contrast, our system faces greater challenges in initial alignment, as close-up stereo images result in less structured scenes and are more susceptible to occlusions and missed detections. Recently, Fusaro \textit{et al.}~\cite{reidentify} presented a system for segmentation and re-identification of strawberries over time. Yet, their transformer-based matching approach is limited to processing one fruit at a time and relies on a greedy algorithm when extended to matching multiple fruit.

Transformer architectures have been utilized for similar applications in agriculture. Vision transformers have been used for weed and crop classification~\cite{transformer_0} and leaf disease detection~\cite{transformer_1}. More recently, transformer-inspired architectures have been used for 3D shape completion~\cite{transformer_shape} and multi-object tracking of fruit~\cite{trans_motdetr}. 


\section{Methodology}
\label{sec:methodology}

\begin{figure*}[t]
    \centering
    \includegraphics[width=\linewidth]{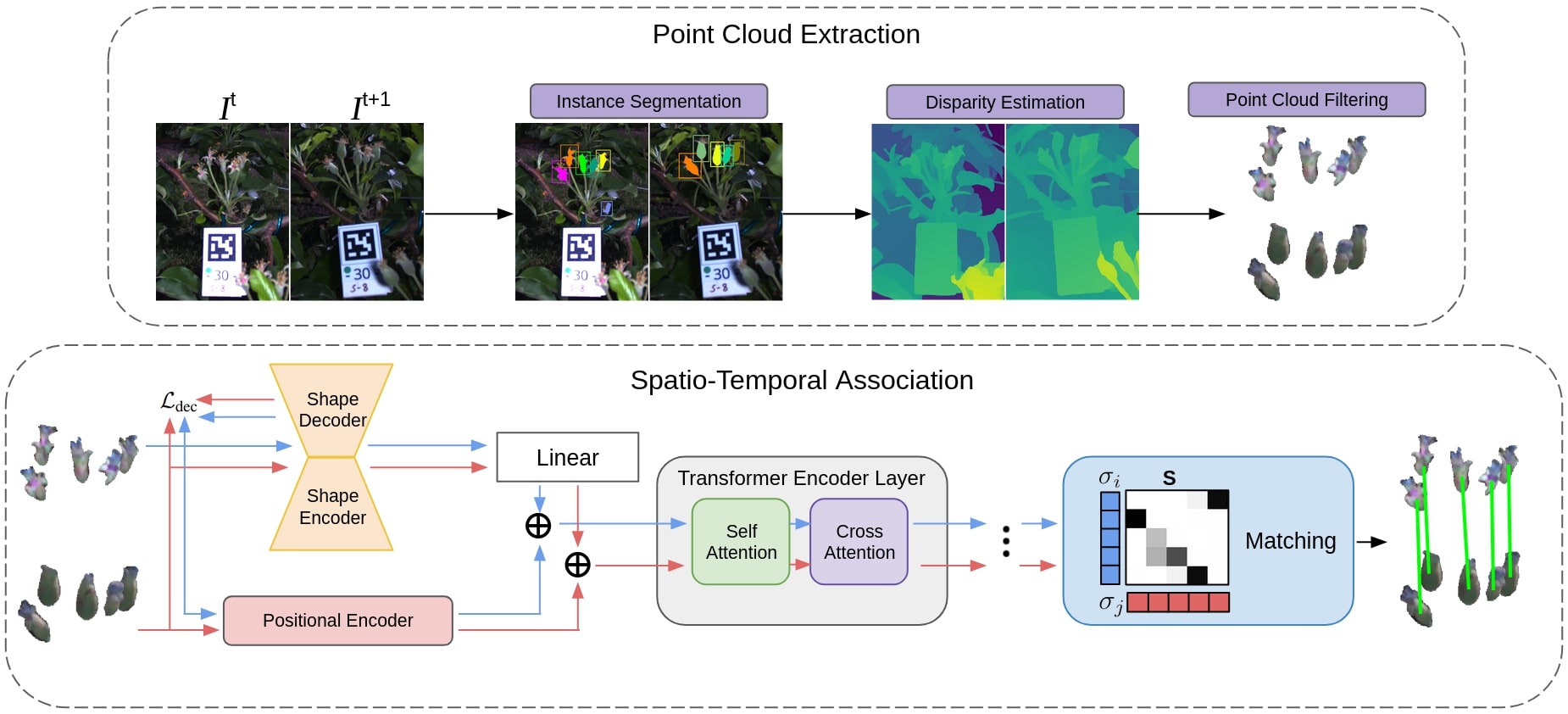}
    \caption{Overview of our point cloud extraction and spatio-temporal association pipeline}
    \label{fig:full_pipeline}
\vspace{-10pt}
\end{figure*}

\subsection{Overview}
Our goal is to establish correspondences between fruitlets in images of the same cluster $C$ taken on different days from varying camera poses. Given two images $I^t$ and $I^{t+1}$ of cluster $C$, let $\mathbf{F}^t = \{f_0^t, \dots, f_{M-1}^t\}$ and $\mathbf{F}^{t+1} = \{f_0^{t+1}, \dots, f_{N-1}^{t+1}\}$ denote the sets of detected fruitlets, with $M$ and $N$ fruitlets respectively. The objective is to determine a set of correspondences $\mathcal{M} = \{(i, j) \mid f_i^t \leftrightarrow f_j^{t+1} \}$ such that each matched pair represents the same fruitlet. Each fruitlet can have at most one correspondence, and some fruitlets may remain unmatched due to occlusions, fruit drop, or missed detections. An AprilTag~\cite{apriltag} is hung next to each cluster to identify the cluster $C$ being imaged (Fig.~\ref{fig:cluster_example}).

An overview of our pipeline is shown in Fig.~\ref{fig:full_pipeline}. On different days, images $I^t$ and $I^{t+1}$ of cluster $C$ are captured using an in-hand version of the flash stereo camera presented by Silwal \textit{et al.}~\cite{stereo_cam}. The fruitlets are segmented using a Mask R-CNN~\cite{mask-rcnn} instance segmentation network, and disparities are estimated with RAFT-Stereo~\cite{lipson2021raft}. For each fruitlet $f_i \in \mathbf{F}^t \cup \mathbf{F}^{t+1}$ belonging to $C$, the corresponding segmentation mask and disparity map are used to construct point cloud $\mathcal{P}_i$, which undergoes multi-stage filtering.

Each point cloud $\mathcal{P}_i$ is used to construct shape descriptor $\mathbf{d}_i$ and 3D positional descriptor $\mathbf{p}_i$ for each fruitlet, which are combined and passed through a multi-layer transformer encoder architecture. The resulting features are then matched based on pairwise similarity.

For the purpose of this work, we assume it is known which detected fruitlets from the Mask R-CNN network belong to cluster $C$. This assumption allows us to focus on temporal fruitlet association rather than the clustering of fruitlets. However, we do not correct any missed detections. In future work, we plan to integrate our fruitlet clustering approach from previous work into a fully automated pipeline~\cite{freeman2023}.

\subsection{Feature Extraction}
\label{sec:feat_extract}
Accurate spatio-temporal association requires reliable spatial reasoning of fruitlets. To enhance accuracy, each fruitlet point cloud $\mathcal{P}_i$ undergoes a multi-stage filtration process. First, a bilateral filter is applied to smooth depth measurements while preserving edge details. Next, a depth discontinuity
filter~\cite{depth_discon}
is used to mitigate artifacts caused by discontinuous depth estimates, which are common in stereo imaging. This is followed by radial outlier removal and a median distance filter to refine the point cloud. Finally, all point clouds from the same image are centered to ensure consistency in spatial alignment.

Once the point clouds are processed, they are used to construct the shape descriptors of the fruitlets. This is achieved using an encoder-decoder architecture based on MinkowskiEngine~\cite{Minkowski, transformer_shape, reidentify}, chosen for its support of sparse 3D convolutions, allowing compatibility with point clouds from various data modalities. Each fruitlet point cloud $\mathcal{P}_i$ is centered, normalized so that its largest principal component has unit length, and voxelized into a sparse volumetric grid $\mathcal{V}_i$ using a fixed voxel size $v$. The encoder extracts a descriptor $\mathbf{d}_i$ that captures the shape of $\mathcal{V}_i$, while the decoder is used to reconstruct $\mathcal{V}_i$ from $\mathbf{d}_i$. This ensures that $\mathbf{d}_i$ preserves the structural geometry of the fruitlet. The decoder serves to assist in pre-training the encoder. This is achieved by encoding and decoding individual fruitlet shapes, guiding the reconstructed shape to closely match the original. After pre-training, the decoder is discarded. We find that pre-training with the decoder improves training stability and speed.

To supervise training, a classifier head is attached after each transposed convolutional layer in the decoder to predict whether each voxel is occupied or not. Voxels classified as unoccupied are pruned before the next layer, ensuring efficient reconstruction and reduced computation. We use a shape descriptor over a visual descriptor to mitigate the challenges posed by inconsistent illumination in agricultural settings and to maintain compatibility with colorless point clouds obtained from other data modalities.

To encode the spatial characteristics of fruitlet $f_i$, we extract a learned positional descriptor $\mathbf{p}_i$ from its point cloud $\mathcal{P}_i$. First, Principal Component Analysis (PCA) is applied to $\mathcal{P}_i$ to determine its principal axes. Using the two main components, we compute the enclosed bounding box in the $xy$ plane. For the $z$ coordinate, we select the median value instead of the extrema, as stereo reconstructions of fruitlets often exhibit flattening along the depth axis as a result of their small size. The selected four keypoints, representing the corners of the bounding box with the median $z$, are then transformed back to the original coordinate space using the inverse PCA transform. 

\begin{equation}
\begin{aligned}
    p_i = \text{MLP} \left( \text{PCA}^{-1} \left( 
    \begin{bmatrix}
        x_-, y_-, z_{\text{med}} \\
        x_{-}, y_{+}, z_{\text{med}} \\
        x_{+}, y_{-}, z_{\text{med}} \\
        x_{+}, y_{+}, z_{\text{med}}
    \end{bmatrix}
    \right) \right), \\[10pt]
    \text{where}
    \begin{bmatrix}
        x_{-}, x_{+} \\ 
        y_{-}, y_{+} \\ 
        z_{\text{med}}
    \end{bmatrix}
    \subset \text{PCA}(\mathcal{P}_i)
\end{aligned}
\end{equation}
Finally, these keypoints are passed through a two-layer MLP to produce the positional descriptor. This simple yet effective method captures both the position and size of the fruitlet, providing essential spatial context for accurate association. The final feature descriptor of fruitlet $f_i$ is given by the combination of its shape and positional descriptors $(\mathbf{d}_i, \mathbf{p}_i)$.

\subsection{Transformer Encoder}
Inspired by state-of-the-art feature-matching networks~\cite{lightglue, loftr}, we employ a multi-layer transformer encoder with alternating self and cross-attention layers to encode the feature descriptors of each fruitlet. First, the shape descriptor $\mathbf{d}_i$ for each fruitlet $f_i$ is passed through a linear layer and summed with the positional descriptor $\mathbf{p}_i$ to form an initial feature vector ${}^{0}\mathbf{x}_i$. 

\begin{equation} 
\label{eq: init_feat}
{}^{0}\mathbf{x}_i = [\text{Linear}(\mathbf{d}_i) + \mathbf{p}_i]
\end{equation} 

Given all initial feature vectors $\mathbf{x}_i \in {}^{0}X^t \cup {}^{0}X^{t+1}$ of fruitlets in $\mathbf{F}^t$ and $\mathbf{F}^{t+1}$, the features are propagated through multiple transformer encoder layers.  Each layer consists of layer normalization, a self-attention block, and a cross-attention block. The self-attention block models spatial relationships between fruitlets in the same image, and the cross-attention block enables matching between fruitlets across different images by modeling the relationship between encoded descriptors. As in ~\cite{transformer}, each self and cross-attention block uses multi-head attention and is followed by a residual connection, layer normalization, a feedforward network, and dropout. The transformer encoder layers refine the feature representations, allowing for robust spatio-temporal association of fruitlets across images.

\subsection{Spatio-Temporal Association}
\label{sec:spatempassoc}
Once the features have been propagated through the transformer encoder layers, the next step is to establish correspondences between fruitlets. To achieve this, the final feature descriptors ${}^L \mathbf{x}_i$ for each fruitlet are passed through a linear layer and its outputs are used to compute a pairwise score matrix $\mathbf{S}$:
\vspace{-12pt}

\begin{equation}
    \mathbf{S_{ij}} = \frac{\text{Linear}({}^L\mathbf{x}_i^t)^\top\text{Linear}({}^L\mathbf{x}_j^{t+1})}{\sqrt{d}} \ \forall (i,j) \in \mathcal{A} \times \mathcal{B}
\end{equation}
where $\mathcal{A} := \{0,...,M-1\}$, $\mathcal{B} := \{0,...,N-1\}$, and $d$ is the feature dimension size. Following \cite{lightglue}, we additionally compute a matchability score $\sigma_i$ for each fruitlet
\vspace{-8pt}

\begin{equation}
    \sigma_i = \text{Sigmoid}(\text{Linear}({}^L\mathbf{x}_i))
\end{equation}
which represents the confidence that a given fruitlet has a valid correspondence in the other image. The score matrix and matchability scores are used to build a soft partial assignment matrix $\mathbf{P}$ (Eq. ~\ref{eq:partial_assign}) using the dual-softmax
operator~\cite{lightglue, loftr}.
Two fruitlets are considered a correspondence if their partial assignment score is the highest among all possible matches in both images and exceeds a predefined threshold.
\vspace{-5pt}
\begin{equation}
\label{eq:partial_assign}
    \mathbf{P}_{ij} = \sigma_i^t\sigma_j^{t+1}\text{Softmax}(\mathbf{S}(i, \cdot))_j\text{Softmax}(\mathbf{S}(\cdot, j))_i
\end{equation}

\subsection{Loss Functions}
To pre-train the shape encoder, we supervise the encoder-decoder network using an auxiliary binary cross-entropy loss at each deconvolution layer, ensuring progressive refinement of voxel occupancy predictions. Given a voxelized fruitlet point cloud $\mathcal{V}_i$, the decoder predicts a voxel occupancy probability ${}^\ell \hat{y}_i$ for each voxel at layer $\ell$, which is compared against the ground truth occupancy label ${}^\ell y_i$ obtained from downsampling $\mathcal{V}_i$ at its corresponding voxel resolution. The decoder loss is formulated as

\begin{equation}
\begin{aligned}
    \mathcal{L}_{\text{aux}}^\ell = -\frac{1}{N_\ell} & \sum_{i=1}^{N_\ell} \left[ {}^\ell y_i \log {}^\ell \hat{y}_i + (1-{}^\ell y_i) \log \left( 1 - {}^\ell \hat{y}_i \right) \right] \\
    &\mathcal{L}_{\text{dec}} = \frac{1}{L} \sum_{\ell=1}^{L} \mathcal{L}_{\text{aux}}^\ell
\end{aligned}
\end{equation}
where $N_l$ is the total number of voxels in layer $\ell$.

To train the transformer encoder layers, we adopt the log-likelihood loss presented in~\cite{lightglue}.
Given ground-truth matches $\mathcal{M}$ and unmatched fruitlets $\overline{\mathcal{A}} \subseteq \mathcal{A}$ and $\overline{\mathcal{B}} \subseteq \mathcal{B}$, the partial assignment matrices and auxiliary losses for each layer are calculated and averaged:

\begin{equation}
\begin{aligned}
\mathcal{L}_{\text{match}} = -\frac{1}{L}& \sum_{\ell=1}^L \left( \frac{1}{|\mathcal{M}|} \sum_{(i,j) \in \mathcal{M}} \log {}^\ell\mathbf{P}_{ij}  \right. \\
&+ \frac{1}{2|\mathcal{\overline{A}}|} \sum_{i \in \mathcal{\overline{A}}} \log \left( 1 - {}^\ell\sigma_i^{t} \right) \\
& \left. + \frac{1}{2|\mathcal{\overline{B}}|} \sum_{j \in \mathcal{\overline{B}}} \log \left( 1 - {}^\ell\sigma_j^{t+1} \right) \right)
\end{aligned}
\vspace{10pt}
\end{equation}

\section{Experiments and Results}

\newcolumntype{A}{>{\centering\arraybackslash\hsize=.25\hsize}X}
\newcolumntype{B}{>{\centering\arraybackslash\hsize=.5\hsize}X}
\newcolumntype{D}{>{\centering\arraybackslash\hsize=.4\hsize}X}
\newcolumntype{E}{>{\centering\arraybackslash\hsize=.1\hsize}X}
\newcolumntype{F}{>{\centering\arraybackslash\hsize=.1\hsize}X}
\newcolumntype{G}{>{\centering\arraybackslash\hsize=.1\hsize}X}
\newcolumntype{H}{>{\centering\arraybackslash\hsize=.1\hsize}X}
\newcolumntype{I}{>{\centering\arraybackslash\hsize=.1\hsize}X}
\newcolumntype{J}{>{\centering\arraybackslash\hsize=.4\hsize}X}
\newcolumntype{K}{>{\centering\arraybackslash\hsize=.1\hsize}X}
\newcolumntype{Q}{>{\centering\arraybackslash\hsize=.05\hsize}X}


\subsection{Dataset}
Our dataset consists of stereo images collected over three years (2021, 2022, and 2023) of Fuji, Gala, and Honeycrisp varietals. The data was collected at the University of Massachusetts Amherst Cold Spring Orchard. 
Image sequences were
collected both by humans using a handheld illumination-invariant flash stereo camera~\cite{stereo_cam, freeman2023} and by a robotic system equipped with the same camera mounted on the end of a 7 DoF robotic arm~\cite{pruning_silwal2021bumblebee, freeman_nbv}. 
A subset of 1350 images were randomly selected and manually labeled.
This resulted in a total of 4,362 cross-day image pairs, with images taken between one and seven days apart, except for six-day intervals, as no capture sessions occurred exactly six days apart in any of the three years of data collection. The distribution of image pairs across different day-apart intervals is approximately 19\%, 38\%, 18\%, 11\%, 8\%, and 6\% for one, two, three, four, five, and seven days respectively.

To achieve proper data separation, the dataset was divided by cluster into training, validation, and test sets following a 60/20/20 split, resulting in 2645, 924, and 793 cross-day image pairs respectively. An additional 600 images from the training clusters were selected to train the Mask R-CNN network, using a 70/15/15 train, validation, and test split.

\subsection{Implementation Details}
Our spatial encoder-decoder operates on a normalized voxel size of 0.03 and follows a hierarchical structure with four downsampling blocks for encoding and four upsampling blocks for decoding. The network is pre-trained for 200 epochs using the ADAM optimizer with a batch size of 64. The initial learning rate is set to $10^{-3}$ and reduces by 90\% every 50 epochs. Our transformer encoder consists of four transformer encoder layers with eight heads used for self and cross attention. It is trained for 200 epochs using the ADAM optimizer. The scheduler includes a warmup phase of 10 epochs, after which the learning rate decays from $10^{-4}$ linearly every training step. For both training procedures, the model achieving the best performance on the validation set is selected for final testing. Additionally, the validation set is used to determine the matching threshold from Sec.~\ref{sec:spatempassoc}.

\subsection{Data Augmentation}

Both networks were trained using standard data augmentation techniques to improve generalization and robustness. To pre-train the spatial encoder, each fruitlet was randomly rotated and subjected to random axis flipping. 
Additionally, the leaf distortion method from~\cite{leaf_distort} was adapted and applied, along with elastic deformation from~\cite{8333411}
to introduce realistic shape variations. Gaussian noise was added to the point clouds as jitter.

To train the transformer encoder, each fruitlet cluster was randomly scaled by $\pm10\%$ and rotated up to 10 degrees. Within each cluster, individual fruitlets underwent an additional random rotation of 10 degrees and were further scaled by $\pm10\%$. Gaussian noise was added as jitter, and each fruitlet’s position was randomly shifted. To introduce variability in point density, points from each fruitlet’s point cloud were randomly dropped.

\begin{table}[t]
\scriptsize
\caption{Results and baseline comparison for average Precision, Recall, and F1-Scores across the entire test set.}
\vspace{-5pt}
\centering
\begin{tabularx}{\linewidth}{@{}AJJJ@{}} 
\hline\hline 
\addlinespace[0.1cm]
\ & Precision (\%) & Recall (\%) & F1-Score (\%)\\
\hline
\addlinespace[0.1cm]
Ours & \textbf{92.9} & \textbf{92.2} & \textbf{92.4}\\
\addlinespace[0.1cm]
ICP-Assoc & 89.0 & 90.1 & 89.5 \\
\addlinespace[0.1cm]
Desc-Assoc & 87.0 & 86.2 & 86.4\\
\addlinespace[0.1cm]
Loftr-Assoc & 79.3 & 68.9 & 72.3\\
\addlinespace[0.1cm]
\hline\hline
\end{tabularx}
\vspace{-15pt}
\label{tab:assoc_res}
\end{table}

\subsection{Comparison Baseline}
We evaluate the effectiveness of our temporal association method described in Sec.~\ref{sec:methodology} and compare it against three baseline approaches. The first baseline aligns the point clouds of all fruitlets in the cluster using Iterative Closest Point (ICP-Assoc). Fruitlets are then associated using the Hungarian algorithm~\cite{hungarian}, which minimizes the pairwise Euclidean distances between transformed centroids. Matches exceeding a predefined distance threshold are discarded.

The second baseline follows the method proposed by Riccardi \textit{et al.}~\cite{riccardi} (Desc-Assoc). Since no open-source implementation is available, we re-implemented the approach based on the details provided in the original paper. The weighting parameters were optimized using Optuna~\cite{optuna_2019} on the training and validation clusters. Three adjustments were made to the original approach. First, we removed the radius term, as fruitlet radii can vary significantly due to growth and partial occlusions. Second, initial point cloud alignment was performed using only ICP and not RANSAC~\cite{ransac}. To improve registration accuracy, ICP was applied on a local volume of the fruitlet cluster rather than the entire scene, which improved registration results. Lastly, instead of selecting a fixed number of nearest neighbors from all fruitlets in the scene, we included all fruitlets within the cluster. This adjustment accounts for the natural variation in cluster sizes  and mitigates the impact of differing camera poses and missed detections.

\begin{table}[t]
\scriptsize
\caption{Ablation comparison for Average Precision, Recall, and F1-Scores across the entire test set.}
\vspace{-5pt}
\centering
\begin{tabularx}{\linewidth}{@{}BDDD@{}} 
\hline\hline 
\addlinespace[0.1cm]
\ & Precision (\%) & Recall (\%) & F1-Score (\%)\\
\hline
\addlinespace[0.1cm]
Ours & \textbf{92.9} & \textbf{92.2} & \textbf{92.4}\\
\addlinespace[0.1cm]
No Shape Desc & 91.2 & 90.6 & 90.8 \\
\addlinespace[0.1cm]
No Pos Desc & 66.3 & 71.1 & 68.3\\
\addlinespace[0.1cm]
No Shape Pre-Train & 90.9 & 90.5 & 90.6\\
\addlinespace[0.1cm]
\hline\hline
\end{tabularx}
\label{tab:ablation_res}
\vspace{-15pt}
\end{table}

For the third baseline, we utilize the Loftr feature matcher from~\cite{loftr} to associate fruitlets based on dense image features (Loftr-Assoc). The images and segmented fruitlet masks are provided as input to the matcher. To determine associations, we use a voting mechanism where candidate fruitlet matches are validated based on the number of feature correspondences within the segmentation masks. We also experimented with the sparse feature matcher from ~\cite{lightglue}, but it produced too few matches within the segmented boundaries.

\subsection{Association Results}

\begin{figure}[!h]
    \centering
\includegraphics[width=\linewidth]{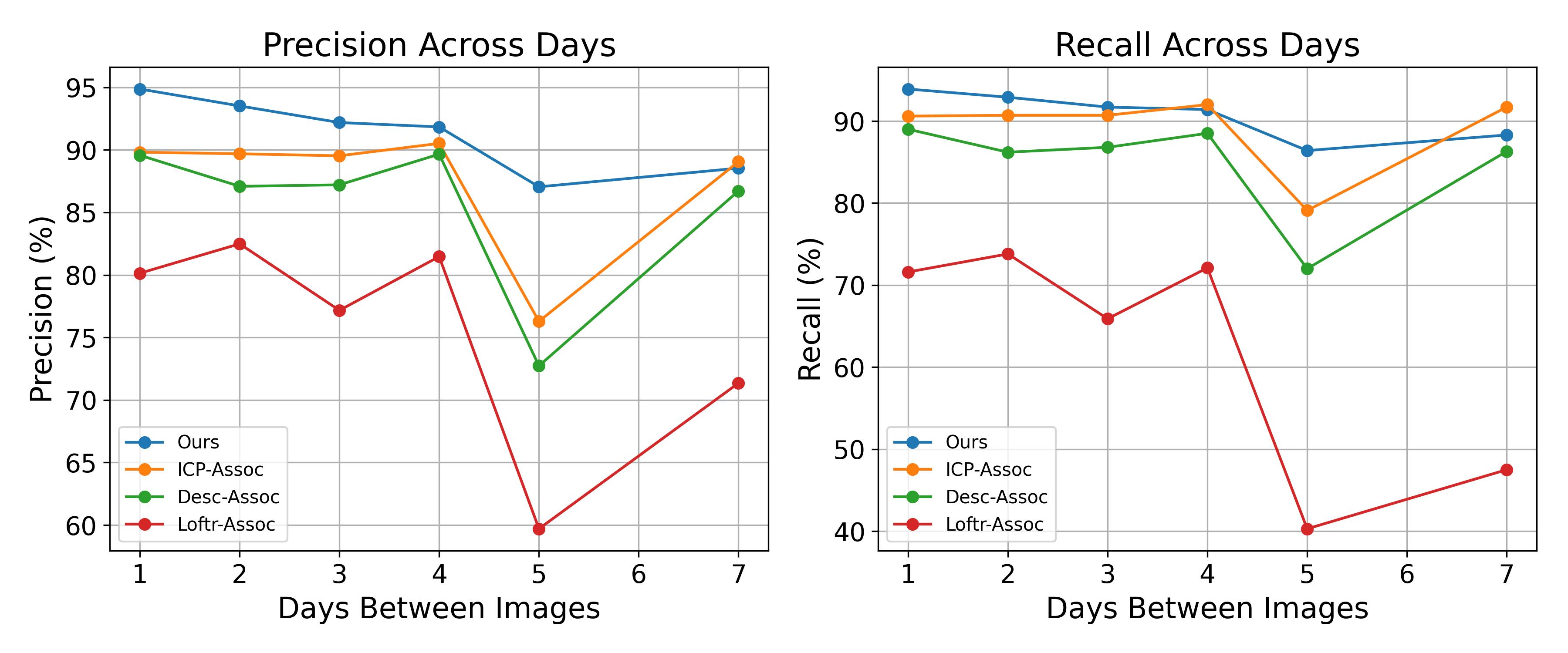}
\vspace{-20pt}
    \caption{Results and baseline comparison for precision and recall when matching is performed across a specified number of days.}
    \label{fig:prec_rec}
\vspace{-10pt}
\end{figure}

\begin{figure*}[t]
    \centering
    \includegraphics[width=\linewidth]{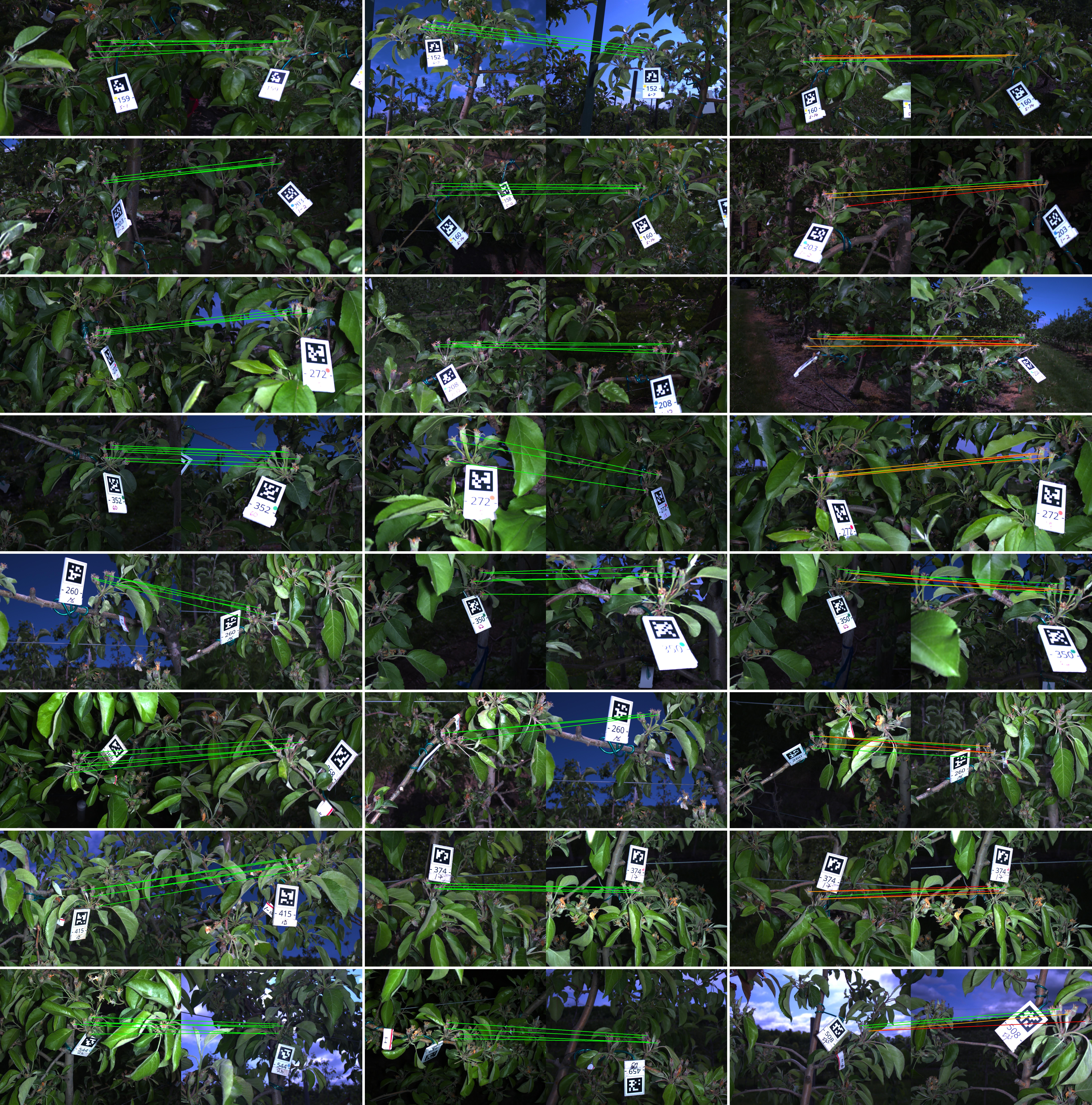}
    \caption{Examples of spatio-temporal association results. Left column: correctly associated fruitlets. Middle column: correctly associated fruitlets when a fruitlet is occluded or fallen off. Right: incorrect association examples where red and orange lines indicate false positive and negative matches respectively.}
    \label{fig:qual_examp}
\vspace{-5pt}
\end{figure*}

In Table~\ref{tab:assoc_res}, we report the average precision, recall, and F1-scores across the entirety of the test set. Our proposed method outperforms all baselines. In particular, we are able to achieve average precision, recall, and F1-scores of $92.9\%$, $92.2\%$, and $92.4\%$ respectively, which is an approximate improvement of 3\% over the next best method. Furthermore, in Fig.~\ref{fig:prec_rec}, we show the precision and recall when matching is performed across a specified number of days between one and seven. Up to five days, our method achieves the highest precision and recall. For image pairs captured seven days apart, our approach has lower recall than ICP-Assoc. This is likely due to the network's difficulty in learning to associate fruitlets with greater shape variation and spatial displacement as a result of fewer training samples for that time interval. Fig.~\ref{fig:qual_examp} presents a qualitative assessment of the performance of our method on selected examples from the dataset.

\subsection{Ablations}
\label{sec:ablation}
We ran ablation tests to evaluate the impact of various design choices on our method. The ablation tests include 
\begin{enumerate}
    \item \textit{No Shape Desc}: The shape descriptor is removed leaving only the positional descriptor.
    \item \textit{No Pos Desc}: The positional descriptor is removed leaving only the shape descriptor.
    \item \textit{No Shape Pre-Train}: The shape encoder is not pre-trained.
\end{enumerate}

\begin{figure}[h]
    \centering
\includegraphics[width=\linewidth]{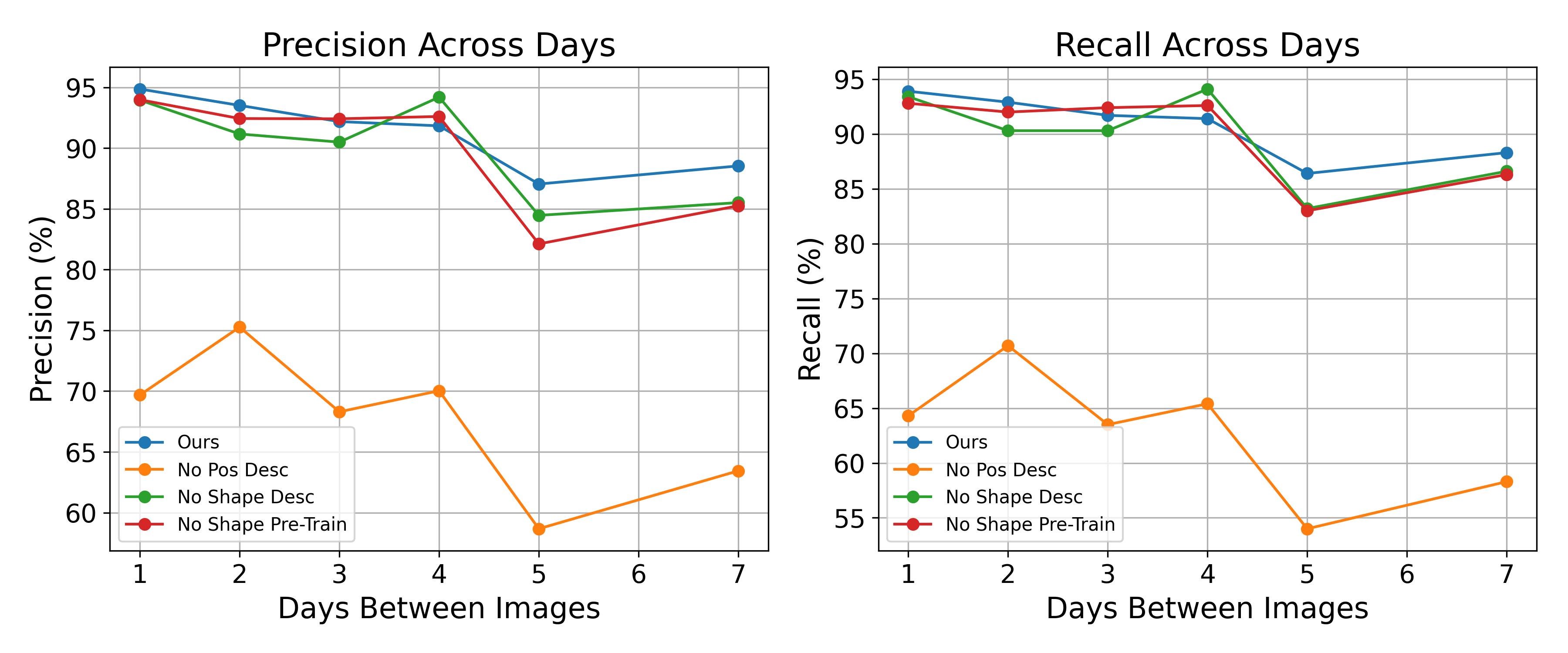}
    \vspace{-20pt}
    \caption{Ablation comparison for precision and recall when matching is performed across a specified number of days.}
    \label{fig:prec_rec_ab}
\vspace{-5pt}
\end{figure}

The average precision, recall, and F1-scores are shown in Table~\ref{tab:ablation_res} and cross-day precision and recall curves in Fig.~\ref{fig:prec_rec_ab}. Our method achieves the highest average precision, recall, and F1-score across all training pairs. Of all components, the positional descriptor has the greatest effect on matching accuracy. This is likely because the similar fruitlet shapes and incomplete point clouds make spatial information essential for accurate matching. The shape encoder and shape encoder pre-training improve the average reported metrics by approximately $1.5\%$, with their impact becoming more pronounced after five days. This may be due to the decreasing reliability of the spatial encoder as fruitlets diverge in shape during growth, making shape encoding increasingly important.

\section{Generalizability to Alternative Datasets}
To evaluate the robustness and generalizability of our approach beyond stereo images of apple fruitlets, we conduct experiments on the Pheno4D~\cite{pheno4d} dataset, which provides temporally segmented point clouds of tomato and maize plants. Unlike our primary dataset, 
Pheno4D captures data using a high-precision laser scanner, resulting in a different modality with higher spatial accuracy and denser point clouds. Additionally, the structural complexity of the Pheno4D dataset differs from our apple fruitlet dataset. While fruitlet clusters contain five to six fruitlets per scene, Pheno4D has denser foliage with up to 50 leaves per plant.

When evaluating our method on the Pheno4D dataset, we used scans separated by two days. 
We performed a 7-fold validation due to the limited number of plants available. Since Pheno4D consists of only seven tomato and maize plants, and transformers typically require large amounts of data, we systematically removed one plant from the training set in each fold and computed the average metrics across all folds. Each fold was trained for 1000 epochs and the checkpoint from the last epoch was used. To account for the fact that the point clouds in Pheno4D are pre-aligned due to the controlled lab setting—unlike in field conditions—we randomly rotated the clouds by ±20 degrees along the z-axis during testing and averaged the results over 10 runs for each fold. Additionally, the positional descriptor described in Sec.~\ref{sec:feat_extract} was modified to use the PCA-aligned bounding box instead of the median $z$ coordinate.

We report the mean and standard deviation of the matching precision, recall, and F1-scores across all folds and testing iterations (Table~\ref{tab:pheno4d_res}), as well as the available results from recent related works~\cite{heiwolt2023statistical, pandey2024spatio}. On the tomato leaf dataset, we are able to achieve mean precision, recall, and F1-scores of $93.8\%$, $91.9\%$, and $92.1\%$ with a standard deviation around $8\%$. For the maize leaves using the leaf tip segmentations, the mean precision, recall, and F1-scores are $96.9\%$, $95.8\%$, and $96.4\%$ with standard deviations around 0.3. While we outperform the available metrics from ~\cite{heiwolt2023statistical, pandey2024spatio}, we recognize that differences in training data, evaluation protocols, and experimental setups may influence direct comparisons. 


\begin{table}[t]
\scriptsize
\caption{Mean and Standard Deviation of Precision, Recall, and F1-Scores across folds and test runs on Pheno4D}
\vspace{-5pt}
\centering
\begin{tabularx}{\linewidth}{@{}lDDD@{}} 
\hline\hline
\addlinespace[0.1cm]
\ & Precision (\%) $\mu/\sigma$ & Recall (\%) \ $\mu/\sigma$ & F1-Score (\%) $\mu/\sigma$\\
\hline
\addlinespace[0.1cm]
\textbf{Tomato Leaf Dataset} & & & \\
Ours & 93.8 \ / \ 7.79 & \textbf{91.9} \ / \ 8.36 & 92.1 \ / \ 7.79\\
Heiwolt \textit{et al.} \cite{heiwolt2023statistical} & -- & 75.4 \ / \ \ -- \ \ \ & -- \\ 
\addlinespace[0.1cm]
\textbf{Maize Dataset} & & & \\
Ours & \textbf{96.9} \ / \ 0.22 & 95.8 \ / \ 0.34 & 96.4 \ / \ 0.28\\
Pandey \textit{et al.} \cite{pandey2024spatio} & 86.2 \ / \ \ -- \ \ \  & -- & -- \\ 
\addlinespace[0.1cm]
\hline\hline
\end{tabularx}
\label{tab:pheno4d_res}
\vspace{-18pt}
\end{table}
\section{Conclusion}
In this work, we present a transformer-based approach for spatio-temporally associating apple fruitlets from stereo images, demonstrating its effectiveness on field data and generalizability across datasets. Our method improves robustness in tracking fruitlets despite occlusions and growth variations. Future improvements include incorporating visual descriptors to enhance association accuracy, although this would come at the expense of applicability to datasets with colorless point clouds. In our future work, we aim to integrate our approach with a next-best-view planning robotic system~\cite{freeman_nbv} to develop a fully automated pipeline for capturing images, clustering, and associating apple fruitlets to ultimately track growth rates and improve yield.

\section*{Acknowledgments}
We thank the University of Massachusetts Amherst Cold Spring Orchard for allowing us to collect data. This research was funded by NSF / USDA NIFA 2020-01469-1022394, NSF / USDA NIFA AIIRA AI Research Institute 2021-67021-035329, and NSF Robust Intelligence 195616.



\bibliographystyle{IEEEtran} 
\bibliography{mybib}

\end{document}